\documentclass[journal]{IEEEtran}
\usepackage{graphicx}
\usepackage{subfig}
\usepackage{amsmath, amssymb, amsfonts}
\usepackage{booktabs}
\usepackage{hyperref}
\usepackage{multirow}
\usepackage{cite}

\usepackage{color}

\begin{document}

% \title{Disentangled Adversarial Autoencoder for Universal Feature Extraction in Physiological Biosignals}
\title{Disentangled Adversarial Autoencoder for Subject-Invariant Physiological Feature Extraction}

\author{Mo Han\textsuperscript{1}, Ozan {\"O}zdenizci\textsuperscript{1}, Ye Wang\textsuperscript{2}, Toshiaki Koike-Akino\textsuperscript{2} and Deniz Erdo{\u{g}}mu{\c{s}}\textsuperscript{1}%

\thanks{\textsuperscript{1}Cognitive Systems Laboratory, Department of Electrical and Computer Engineering, Northeastern University, Boston, MA 02115, USA, E-mail: \{han, oozdenizci, erdogmus\}@ece.neu.edu.}
\thanks{\textsuperscript{2}Mitsubishi Electric Research Laboratories (MERL), Cambridge, MA 02139, USA, E-mail: \{yewang, koike\}@merl.com.}%
\thanks{M.~Han was an intern at MERL during this work. O.~{\"O}zdenizci and D.~Erdo{\u{g}}mu{\c{s}} are partially supported by NSF (IIS-1149570, CNS-1544895, IIS-1715858), DHHS (90RE5017-02-01), and NIH (R01DC009834).}
}

\markboth{IEEE Signal Processing Letters}
{Han \MakeLowercase{\textit{et al.}}: Disentangled Adversarial Autoencoder for Subject-Invariant Physiological Feature Extraction}
\maketitle

\begin{abstract}Recent developments in biosignal processing have enabled users to exploit their physiological status for manipulating devices in a reliable and safe manner.
One major challenge of physiological sensing lies in the variability of biosignals across different users and tasks.
To address this issue, we propose an adversarial feature extractor for transfer learning to exploit disentangled universal representations.
We consider the trade-off between task-relevant features and user-discriminative information by introducing additional adversary and nuisance networks in order to manipulate the latent representations such that the learned feature extractor is applicable to unknown users and various tasks.
Results on cross-subject transfer evaluations exhibit the benefits of the proposed framework, with up to $8.8\%$ improvement in average accuracy of classification, and demonstrate adaptability to a broader range of subjects.\end{abstract}


\IEEEpeerreviewmaketitle

\section{Introduction}
\label{sec:intro}

\IEEEPARstart{R}{ecently}, biosignal processing has obtained increasing significance, since the abilities of machines to understand human emotions, discern physiological disorders, and execute appropriate actions are key points in the area of human computer interaction (HCI)~\cite{physiological1-HCI}.
HCI enables users to communicate their physiological information with machines for help with manipulating external devices in a more reliable, robust and safe manner. Traditionally, the assessment of physiological activity (e.g., human stress level and mental status) was implemented by monitoring signals such as electroencephalography (EEG)~\cite{petrantonakis2009emotion} and electromyography (EMG)~\cite{EMG-HANDS-CSL}. %and electrocardiogram (ECG)~\cite{ECG}.
However, these measurements require either surface (non-invasive) or implanted (invasive) electrodes and frequent calibration, which increase system cost and decrease user comfort.
To realize more portable interfaces, significant progress was recently achieved with wearable sensors for precisely monitoring physiological signals such as heart rate, skin temperature, and arterial oxygen level~\cite{NonEEG1,NonEEG2,NonEEG3,giakoumis2013subject,giannakakis2019review,ozdenizci2018time}.
These more convenient (non-EEG) biosignals avoid the aforementioned issues, and can be obtained from a wrist-worn platform in more effective, comfortable, and less expensive ways.

However, biosignals often vary across subjects and recording sessions of the same person depending on physical/mental conditions or the disturbance by task-irrelevant activity.
Such variability is an obstacle to successful HCI applications to a wider range of users and tasks, since biosignals are often collected from limited number of subjects.
Under such restrictions, a robust feature extractor can be constructed with transfer learning~\cite{trans2,trans3,trans4}, which tries to discover shared data features that are invariant across subjects and tasks. 
In particular, promising results were demonstrated for transfer learning by censoring nuisance features via adversarial training~\cite{ozdenizci2020learning, advTrans-csl2, advTrans-csl1, advTrans4,  adversarialClsf, wu2020modality, sun2020adversarial}.
These works use adversarial methods to learn universal features shared by an attribute group, where a discriminative unit distinguishes shared features with respect to the different attributes adversarially to the feature extractor.
% For example in \cite{ozdenizci2020learning, advTrans-csl2,advTrans-csl1} subject-invariant features are learned from EEG, in \cite{advTrans4} cross-face-attribute representations are extracted from images, and \cite{wu2020modality, sun2020adversarial} addressed cross-modal retrieval and cross-project defect prediction.\todo{I feel the previous sentence is lengthy and useless.}
However, in existing works, the adversarial unit will act directly on the entire latent representation to preserve cross-attribute shared features, leading to loss of attribute-specific information.
Hence, instead of simply reserving shared features only with one adversarial discriminator, we disentangle the physiological latent representations into 2 parts of subject- and task-relevant features by jointly training two discriminators, so that the model can better handle both subject- and task-specific variations.
% In \cite{wu2020modality}, the authors use adversarial method to learn modality-shared features for cross-modal retrieval, where a discriminative model distinguishes modality-shared features with respect to the two modalities adversarially to the generative model. Similarly, Sun et al. \cite{sun2020adversarial} explore cross-project defect representations to better learn the data features from source and target projects by adversarially training the feature transformer and project discriminator. Similarly, Sun et al. \cite{sun2020adversarial} explore cross-project defect representations to better learn the data features from source and target projects by adversarially training the feature transformer and project discriminator. 

This paper proposes an extended adversarial feature encoding to exploit disentangled universal representations, motivated by \cite{adversarialClsf}, where the adversarial classifier is generalized into a feature extractor. 
Unlike classic feature extractors ignoring the target subjects and task calibration, we introduce two additional networks, i.e., adversary and nuisance blocks, in an autoencoder (AE) architecture to re-organize the latent representations, thereby accounting for a trade-off between task-related features and person-discriminative information.
Even if a new user is dissimilar to any of the training subjects, instead of reconciling to suboptimality by extracting subject-specific features only, task-relevant representations can still be incorporated into the feature extraction.
Empirical assessments were performed on a publicly available dataset of physiological biosignals for human stress level assessment.
%Performances of the traditional adversarial method and our work are compared in the evaluation.
Results show a significant advantage of the disentangled adversarial framework through cross-subject evaluations with various classifiers, achieving up to $8.8\%$ improvement in classification accuracy. 
%Moreover, it is worth noting that the proposed transfer learning framework is also applicable to other feature extractors and encoders that are available, depending on the signal characteristics but not limited to physiological biosignals only.

%%%%%%%%%%%%%%%%%%%%%%%%%%%%%%%%%%%%%%%%%%%%%%%%%%%%%%%%%%%%%%%%%%%%
\section{Methods}

\subsection{Disentangled Adversarial Transfer Learning}

\begin{figure}[t]
\centering
  \subfloat
  %\subfigure
  [Disentangled feature learning]{\includegraphics[width=\linewidth]{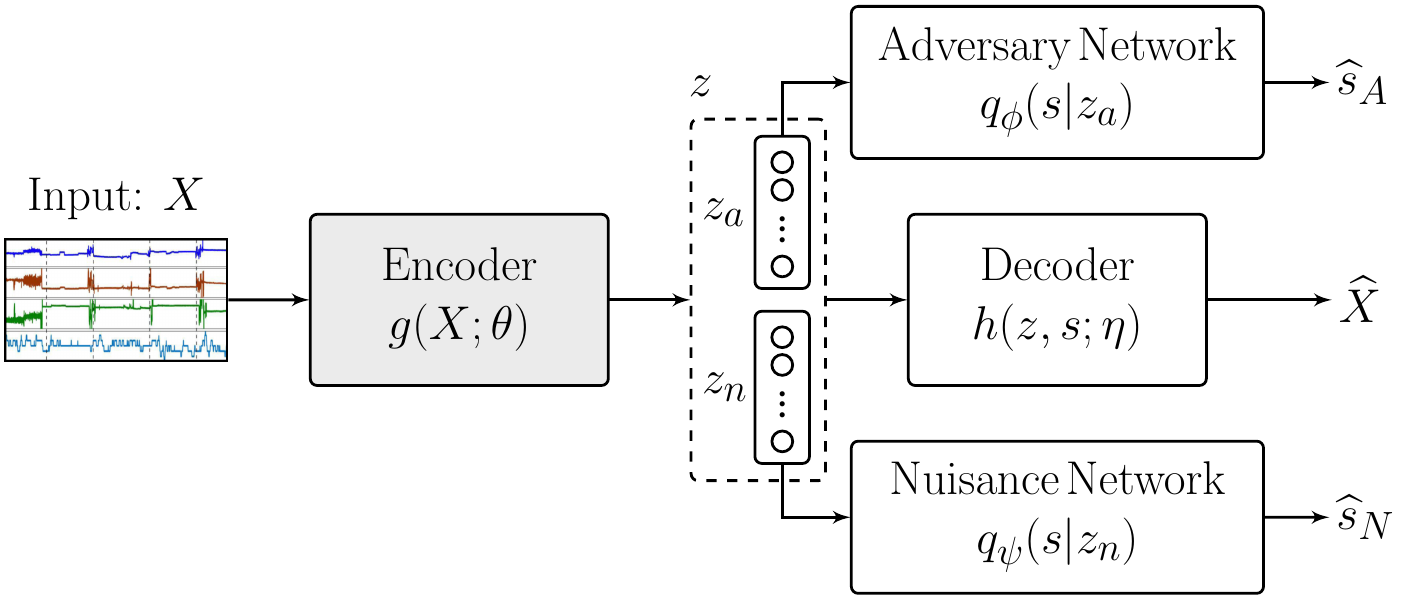}}\\
  \subfloat
  %\subfigure
  [Task-specific classifier training with frozen encoder]{\includegraphics[clip,trim=0 0 0 0.2cm,width=\linewidth]{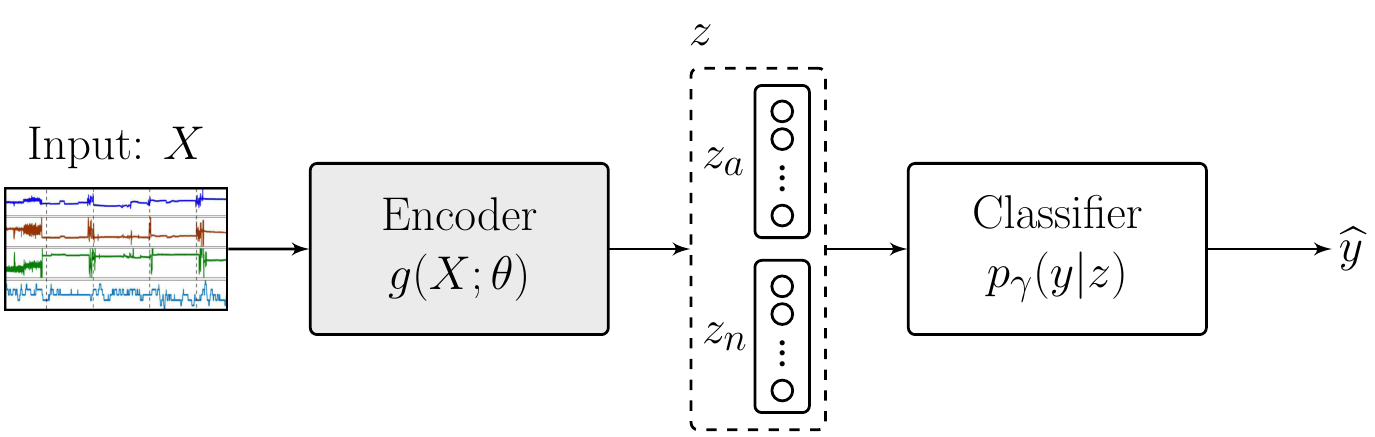}}%\vspace{-0.1cm}
  \caption{(a) An encoder $z=g(X;\theta)$ and decoder $\hat{X}=h(z,s;\eta)$ conditioned on $s$ is trained to learn a latent variable $z$ from data $X$.
  Latent $z$ is divided to $z_a$ and $z_n$, where $z_a$ is input to an \textit{adversary network}, while $z_n$ is fed to a \textit{nuisance network}. 
  The full latent vector $z=[z_a,z_n]$ is used as an input to the decoder, alongside the condition $s$.
  (b) Using the pre-trained encoder frozen as a static feature extractor to generate $z$ form $X$, a classifier is then optimized to predict the corresponding user physiological status or any other task.}
\label{model}
\end{figure}

Define $\{(X_i,y_i,s_i)\} _{i=1}^{n}$ as a training set, where $X_i \in \mathbb{R}^C$ is the data matrix of trial $i$ recorded from $C$ channels, $y_i \in {\{0,1,\ldots,L-1\}}$ is the label of  physiological status/task among $L$ classes, and $s_i \in {\{1,2,\ldots,S\}} $ is the identification (ID) number among $S$ subjects.
We assume the label $y$ and subject ID $s$ are marginally independent, and the data is dependent on $y$ and $s$, i.e., $X\sim p(X|y,s)$.
Our goal is to build a model that estimates the label $y$ of a given observation $X$, where the model is robust to the variability across subjects $s$, which captures the nuisance variations we wish to suppress for transferring feature extraction.
In the proposed method, we first train a disentangled feature extractor based on a modified AE (i.e., an encoder-decoder pair), and then utilize this learned encoder as a static feature extractor to further train a task classifier for the final discriminative model as shown in Fig.~\ref{model}.

% \begin{figure*}[t]
% \centering
%   \subfigure[Disentangled feature learning]{\includegraphics[width=0.5\textwidth]{flowchart.pdf}}
%   \subfigure[Task-specific classifier training with frozen encoder]{\includegraphics[width=0.47\textwidth]{flowchart2.pdf}}\vspace{-0.1cm}
%   \caption{(a) A modified AE with encoder $z=g(X;\theta)$ and decoder $\hat{X}=h(z,s;\eta)$ conditioned on $s$ is trained to learn a latent representation $z$ from input data $X$.
%   Latent $z$ is divided to $z_a$ and $z_n$, where $z_a$ is input to an \textit{adversary network}, while $z_n$ is fed to a \textit{nuisance network}.
%   The full latent vector $z=[z_a,z_n]$ is used as an input to the decoder, alongside the condition $s$. (b) Using the pre-trained encoder $g(X;\theta)$ from part (a) frozen as a static feature extractor, a classifier is then optimized using the encoder output representation $z$ to predict the corresponding user physiological status or any other task.}
% \label{model}
% \end{figure*}

AEs are feature learning machines which constitute an encoder and decoder network pair. The encoder learns a latent vector to represent data features, while the decoder aims to recover the input data from this learned latent representation. Here, we train a modified AE to extract the latent feature $z$ from data $X$ via the encoder $z=g(X;\theta)$ with parameters $\theta$, attached with a decoder $\hat{X}=h(z,s;\eta)$ parameterized by $\eta$, where the decoder is conditioned on the nuisance variable $s$ as an additional input along with $z$. The conditional decoder output $\hat{X} \in \mathbb{R}^C$ is a reconstructed estimate of input $X$.

In the proposed model, the latent representation $z$ consists of two sub-parts: $z_a$ and $z_n$, divided by the ratio of $(1-r_N):r_N$ over their dimensionality.
The representation $z_a$ is fed into an \textit{adversary network} with parameters $\phi$, while the feature $z_n$ is input to another \textit{nuisance network} parameterized by $\psi$, as illustrated in Fig.~\ref{model}(a).
The full latent space $z=[z_a,z_n]$  is further fed into the decoder $h(z,s;\eta)$, which is conditioned on $s$. 
Disentangling $z$ into sub-parts $z_a$ and $z_n$ is proposed to systematically re-arrange the features related to task and subject respectively: $z_a$ aims to conceal the subject information regarding $s$, while $z_n$ is designed to include the subject-related features.
By dissociating the nuisance variable from task-related features, the model is extrapolated into a broader domain of subjects.
For an unknown user, the subject-invariant feature $z_a$ would be useful for the final prediction; simultaneously, the biosignal which is similar to known subjects can also be projected to $z_n$ as an additional side information. 
In order to embed more task-related information into $z_a$ and filter out factors of variation caused by $s$, the encoder is forced to minimize the likelihood $q_\phi\left(s|z_a\right)$ of $z_a$; on the other hand, to retain sufficient subject-related information within $z_n$, the encoder is simultaneously designed to maximize the likelihood $q_\psi\left(s|z_n\right)$ from $z_n$.
%The full representation $z$ is fed to the decoder, which is conditioned on $s$ as an additional input besides $z$. 
The encoder-decoder pair is trained to minimize the reconstruction loss between $X$ and $\hat{X}$. 
Hence, the overall loss to train the proposed model is given by
\begin{align}
\label{DA-cAEloss}
         &\mathsf{Loss}_\mathrm{AE}(X;\eta,\theta,\psi,\phi) =
           - \mathbb{E} \big[ \log p_\eta \big(\hat{X}|g(X;\theta), s \big) \big]  \notag\\
          & \quad {} - \lambda_N \mathbb{E} \left[ \log q_\psi \left(s|z_n \right) \right]   
           + \lambda_A \mathbb{E} \left[ \log q_\phi \left(s|z_a \right) \right],
\end{align}
where the first term is the reconstruction loss of the decoder $\hat{X}=h(z,s;\eta)$ with $z=g(X;\theta)$, and $\lambda_A$ and $\lambda_N$ denote the weight parameters for adversary and nuisance networks respectively, to implement an adjustable trade-off between invariance and identification performance. When $\lambda_A = \lambda_N = 0$, the model reduces to a regular conditional AE (cAE) structure without the disentangling transfer learning units. 

In addition to the overall objective, at each training iteration, the adversary and nuisance networks are optimized towards predicting the variable $s$ among $S$ subjects by maximizing the likelihoods $q_\phi\left(s|z_a\right)$ and $q_\psi\left(s|z_n\right)$ respectively. For the parameter updates at each iteration, optimization is performed by stochastic gradient descent alternatingly among the adversary network, nuisance network and encoder-decoder pair, where the adversary and nuisance networks are individually trained to minimize their cross-entropy losses.

Attached to the pre-trained disentangled encoder whose network weights are frozen, a separate classifier is then trained using the feature representation $z$, as shown in Fig.~\ref{model}(b). The task classifier aims to predict the user physiological status or task category $y$ given observation $X$ among $L$ classes, where $X$ would pass through the feature extractor $z=g(X;\theta)$ before the task classifier. Classifier training is performed to minimize the softmax cross-entropy loss:
%\begin{equation}
%\label{clsloss}
$
%          \mathsf{Loss}_{C}(z;\gamma) =
          \mathbb{E} \left[ -\log p_\gamma \left(\hat{y}|z \right) \right]
          $,
%\end{equation}
where $\hat{y}$ is the task category estimate, and $\gamma$ are task classifier parameters.

\subsection{Model Architecture}

Deep learning frameworks have shown promising performance in biosignal processing recently \cite{advTrans-csl1,atzori2016deep,faust2018deep}. In the light of these works, we mainly focus on neural network feature extractor. We however note that any other discriminative learning methods can be used in the proposed methodology of disentangled adversarial transfer learning. The model architecture specifications we used in our experiments are presented in Table~\ref{model_architectures}. Latent representation $z$ with dimensionality $d=15$ is generated and split into sub-representations $z_n$ and $z_a$ with dimensions of $d \cdot r_N$ and $d \cdot (1 - r_N)$, which are respectively fed into adversary and nuisance networks with the same output dimensionality $S$ for the classification of subject IDs.

% The encoder and decoder both consist of two linear layers with $15$ nodes per layer. We note that we did not observe significant improvements by deepening the network or altering the number of units for our physiological biosignal dataset under test. Representation $z$ with dimension $d=15$ is then generated and split into $z_n$ and $z_a$ with dimensions of $d \cdot r_N$ and $d \cdot (1 - r_N)$ respectively. The adversary network and nuisance network are constructed as a multilayer perceptron (MLP) with a single hidden layer and rectified linear unit activation. Sub-representations $z_a$ and $z_n$ are respectively fed into adversary and nuisance networks with same output dimension $S$ for the classification of subject IDs. %Complete representation $z$ is then utilized as the input to the decoder with an output dimensionality of $C \times T$ for signal reconstruction. 

To verify the adaptability of the proposed feature extractor framework with adversarial disentangling, multiple structures for task classification were implemented, including multilayer perceptron (MLP), nearest neighbors, decision tree, linear discriminant analysis (LDA), linear support vector machine (SVM), and logistic regression classifiers. % whose output dimensions are $L$ for task category classification.

% \begin{table}[t]
%     \caption{Network architectures (ReLU: rectified linear unit).}
%     \centering
%     \resizebox{0.39\textwidth}{11.5mm}{
%     \begin{tabular}{ c | c| c  } 
%     \hline
%     Layer & Input \& Output Dim. & Operation \\ 
%     \Xhline{2.5\arrayrulewidth}
%     Encoder-1 & $C \times T$ $\rightarrow$ $15 \times 1$  & Linear layer + ReLU \\\hline
%     Encoder-2 & $15 \times 1$ $\rightarrow$ $15 \times 1$  & Linear layer + ReLU \\\hline
% %    Latent ($z$) & $d=15$  &  \\\hline
%     Decoder-1 & $15 \times 1$ $\rightarrow$ $15 \times 1$  & Linear layer + ReLU \\\hline
%     Decoder-2 & $15 \times 1$ $\rightarrow$ $C \times T$  & Linear layer + ReLU \\
%     \Xhline{2.5\arrayrulewidth}
%     Adversary & $15 \times 1$ $\rightarrow$ $S \times 1$  & Single linear layer \\\hline
%     Nuisance & $15 \times 1$ $\rightarrow$ $S \times 1$  & Single linear layer \\\hline
%     \end{tabular}
%     }\vspace{-0.1cm}
%     \label{model_architectures}
% \end{table}

\begin{table}[t]
    \caption{Network architectures; FC($d_\mathrm{i}, d_\mathrm{o}$): fully connected linear layer with input/output dimensions $d_\mathrm{i}$ and $d_\mathrm{o}$, ReLU: rectified linear unit}
    \centering
    %\scalebox{1.15}{
    \begin{tabular}{ c  c  } 
    %\hline
    % & Layer(input dim., output dim.)  \\ 
    \toprule %Xhline{2.5\arrayrulewidth}
    Encoder Network & FC($C$, 15) $\rightarrow$ ReLU $\rightarrow$ FC(15, 15)\\\hline
    Decoder Network & FC(15, 15) $\rightarrow$ ReLU $\rightarrow$ FC(15, $C$) \\
    \hline %Xhline{2.5\arrayrulewidth}
    Adversary Network  & FC(15, $S$) \\\hline
    Nuisance Network & FC(15, $S$) \\\bottomrule
    \end{tabular}
    %}
    %\vspace{-0.1cm}
    \label{model_architectures}
\end{table}

%%%%%%%%%%%%%%%%%%%%%%%%%%%%%%%%%%%%%%%%%%%%%%%%%%%%%%%%%%%%%%%%%%%%
\section{Experimental Evaluation}
% our aim: keep main acc stable while decrease adv acc and increase nui acc

\subsection{Physiological Biosignal Dataset}

We evaluate our model on a publicly available biosignal dataset for assessment of stress status levels~\cite{NonEEG1}.
This database consists of multi-modal physiological biosignals for inferring $L=4$ discrete stress status levels from $S=20$ healthy subjects, including physical stress, cognitive stress, emotional stress and relaxation.
The data were collected by non-invasive wrist-worn biosensors, measuring electrodermal activity, temperature, three-dimensional acceleration, heart rate, and arterial oxygen level.
Thus, the data consist of signals from $C=7$ channels in total, which were temporally downsampled to $1$~Hz sampling rate to align all data channels.
%For each stress status, a corresponding task of $35$ minutes long (i.e., $T=2100$ time samples) was assigned to subjects.
Each subject performed $7$ trials, where $4$ out of the $7$ trials were for the relaxation status, over a measurement session lasting approximately $35$ minutes.
To account for imbalanced number of trials across classes, we excluded the last three relaxation trials, resulting in one trial per stress status.

\subsection{Experiment Implementation}

\begin{figure*}[t!]
  \centering
  \includegraphics[clip,trim=0.9cm 0 0.9cm 0.4cm,width=\textwidth]{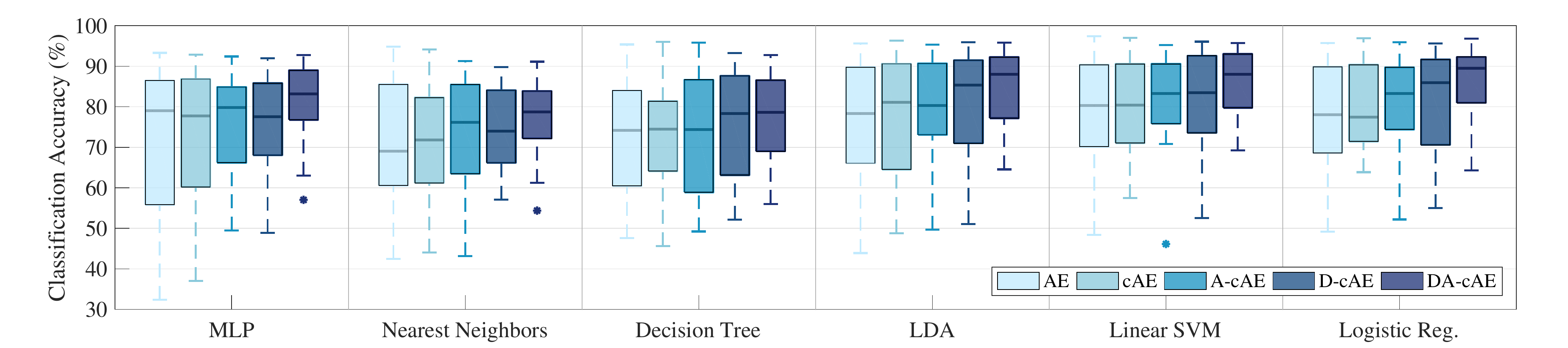}%\vspace{-0.1cm}
  \caption{Transfer learning accuracies for $20$ held-out subjects: (1) regular AE with decoder $h(z;\eta)$, (2) cAE: AE with $s$-conditional decoder $h(z,s;\eta)$, (3) A-cAE: cAE with an extra adversary network, (4) D-cAE: cAE with an extra nuisance network, (5) DA-cAE: cAE with both adversary and nuisance networks. For each box, the central line marks the median, upper and lower bounds represent first and third quartiles, and dashed lines denote extreme values.}%\vspace{-0.1cm}
\label{results_box}
\end{figure*}

The regularization parameters $\lambda_A$ and $\lambda_N$, as well as the ratio of nuisance feature $r_N$ were to be determined. The model was trained with different parameter combinations, and favored decreasing in adversary accuracy with increasing nuisance accuracy, while achieving high task classifier accuracy on validation sets. To narrow down the amount of $\lambda_A$ and $\lambda_N$ parameter combination options, we first set $\lambda_A=0$ to optimize $\lambda_N$; then froze $\lambda_N$ at its optimized value from the previous step to choose an optimal $\lambda_A$. The value ranges we used for these parameters are $\lambda_A \in \{0, 0.01, 0.1, 0.2, 0.5 \}$ and $\lambda_N \in \{ 0, 0.005, 0.01, 0.2, 0.5 \}$. Based on our model that the user-related features $z_n$ will not vary dramatically across different users, we fixed the ratio of nuisance representation to $r_N=1/3$. Note that these parameter combinations can be further optimized by cross-validating the model learning process. Evaluations were performed by cross-subjects transfer analysis using a leave-one-subject-out approach, where the left-out subject constituted the cross-subject test set, and the training and validation sets were composed of $90\%$ and $10\%$ randomized trial splits from the remaining subjects.

\subsection{Results and Discussion}
\label{sec:result}

%In order to evaluate the transfer learning capability of the proposed model structure, we implemented cross-subjects validation with data of one subject held-out each time and repeated the same validation for all subjects one by one. Experiments were executed on various classifiers, whose inputs were representations from the feature extractor with adversarial disentangling units. These classifiers included Multilayer Perceptron (MLP), Nearest Neighbors, Linear SVM, Decision Tree, Logistic Regression, and Linear Discriminant Analysis (LDA), whose output dimensions were $4$ for classifying the stress status. 

Transfer learning accuracies for $20$ held-out subjects with different classifiers and feature learning models are presented in Fig.~\ref{results_box}.
Specifically, AE is a baseline encoder-decoder pair whose decoder is $h(z;\eta)$, cAE is a conditional AE whose decoder $h(z,s;\eta)$ is conditioned on $s$, A-cAE and D-cAE are cAE with only an adversary or nuisance network present respectively, and DA-cAE specifies our cAE with both adversary and nuisance networks.
Corresponding parameter choices for each classifier with each model in Fig.~\ref{results_box} are presented in Table~\ref{params}, which were optimized via a parameter sweep as previously described. 
Note that the A-cAE corresponds to the adversarial learning methods presented in \cite{advTrans-csl2, wu2020modality, sun2020adversarial, advTrans4}.

As observed in Fig.~\ref{results_box} and Table~\ref{params}, when we compare cAE and AE models, simply providing the conditional input $s$ to the decoder can slightly improve the classification performance.
We further observe increased accuracies with A-cAE and D-cAE models compared to cAE, indicating that more subject-shared information remaining in $z_a$ results in better decoding of $y$. 
More importantly, DA-cAE achieves a further improvement of up to $8.8\%$ in an average accuracy compared to the regular AE, and also outperforms individual regularization approaches A-cAE and D-cAE.
With both adversary and nuisance networks, our feature extractor leads to more stable performance universally across all subjects and all task classifiers.
In addition, the worst-case transfer accuracies are highly improved as observed in Fig.~\ref{results_box}, showing that the proposed transfer learning framework yields better robustness to novel users from a broader range by disentangling subject- and task-related representations at the feature extractor end.

% %\renewcommand{\arraystretch}{1.2}
\begin{table*}[t!]
    \caption{Optimized parameters and corresponding averaged cross-subject accuracies on five models for six classifiers}
    \centering
    \resizebox{\textwidth}{13.5mm}{
    \begin{tabular}{c c c c c c c c c c c c c c c c c c c c c}
    \toprule
     & \multicolumn{4}{c}{AE}  & \multicolumn{4}{c}{cAE}  & \multicolumn{4}{c}{A-cAE~\cite{advTrans-csl2, sun2020adversarial, advTrans4, wu2020modality}} & \multicolumn{4}{c}{D-cAE} & \multicolumn{4}{c}{\textbf{DA-cAE (Proposed)}} \\
     \cmidrule(r){2-5} \cmidrule(r){6-9} \cmidrule(r){10-13} \cmidrule(r){14-17} \cmidrule(r){18-21}
     & $\lambda_A$ & $\lambda_N$ & $r_N$ & avg acc & $\lambda_A$ & $\lambda_N$ & $r_N$ & avg acc & $\lambda_A$ & $\lambda_N$ & $r_N$ & avg acc & $\lambda_A$ & $\lambda_N$ & $r_N$ & avg acc & $\lambda_A$ & $\lambda_N$ & $r_N$ & avg acc\\ \toprule
    \textbf{MLP} & 0 & 0 & 0 & \multicolumn{1}{c|}{72.2\%} & 0 & 0 & 0 & \multicolumn{1}{c|}{72.9\%} & 0.005 & 0 & 0 & \multicolumn{1}{c|}{75.0\%} & 0 & 0.005 & 1/3 & \multicolumn{1}{c|}{75.2\%} & 0.01 & 0.005 & 1/3 & \textbf{81.0\%} \\  
    Nearest Neighbors & 0 & 0 & 0 & \multicolumn{1}{c|}{71.1\%} & 0 & 0 & 0 & \multicolumn{1}{c|}{72.2\%} & 0.1 & 0 & 0 & \multicolumn{1}{c|}{73.9\%} & 0 & 0.01 & 1/3 & \multicolumn{1}{c|}{74.9\%} &  0.1 & 0.01 & 1/3 & \textbf{77.0\%} \\  
    Decision Tree & 0 & 0 & 0 & \multicolumn{1}{c|}{71.2\%} & 0 & 0 & 0 & \multicolumn{1}{c|}{72.4\%} & 0.1 & 0 & 0 & \multicolumn{1}{c|}{73.4\%} & 0 & 0.01 & 1/3 & \multicolumn{1}{c|}{75.8\%} & 0.2 & 0.01 & 1/3 & \textbf{77.3\%} \\  
    LDA & 0 & 0 & 0 & \multicolumn{1}{c|}{76.5\%} & 0 & 0 & 0 & \multicolumn{1}{c|}{77.8\%} & 0.05 & 0 & 0 & \multicolumn{1}{c|}{79.8\%} & 0 & 0.2 & 1/3 & \multicolumn{1}{c|}{80.2\%} & 0.2 & 0.2 & 1/3 & \textbf{84.3\%} \\ 
    Linear SVM & 0 & 0 & 0 & \multicolumn{1}{c|}{79.6\%} & 0 & 0 & 0 & \multicolumn{1}{c|}{80.2\%} & 0.005 & 0 & 0 & \multicolumn{1}{c|}{81.6\%} & 0 & 0.005 & 1/3 & \multicolumn{1}{c|}{81.3\%} &  0.2 & 0.005 & 1/3  & \textbf{85.5\%} \\ 
    Logistic Regression & 0 & 0 & 0 & \multicolumn{1}{c|}{78.7\%} & 0 & 0 & 0 & \multicolumn{1}{c|}{79.7\%} & 0.05 & 0 & 0 & \multicolumn{1}{c|}{80.8\%} &  0 & 0.2 & 1/3 & \multicolumn{1}{c|}{81.8\%} & 0.2 & 0.2 & 1/3 & \textbf{85.3\%} \\ 
    \bottomrule
    \end{tabular}
    }
    \label{params}
\end{table*}

\begin{table}[t]
    \caption{Parameter impact on accuracy for MLP classifier %Accuracies for the classifier, adversary and nuisance networks are present.
    }
    \centering
    \begin{tabular}{c c c c c c c}
    % \toprule
    % & $\lambda_A$ & $\lambda_N$ & $r_N$ & \textbf{Main Classifier} & \textbf{Adversary Network} & \textbf{Nuisance Network} \\ \toprule
    & $\lambda_A$ & $\lambda_N$ & $r_N$ & \textbf{MLP} & \textbf{Adversary} & \textbf{Nuisance} \\ \toprule
    AE & 0 & 0 & 0 & 72.2\% & 7.8\% & 5.6\% \\  \midrule
    cAE & 0 & 0 & 0 & 72.9\% & 8.5\% & 5.8\% \\  \midrule
    \multirow{4}{*}{D-cAE} & \textbf{0} & \textbf{0.005} & \textbf{1/3} & \textbf{75.2\%} & \textbf{8.6\%} & \textbf{18.8\%} \\
    & 0 & 0.01 & 1/3 & 74.1\% & 12.1\% & 24.2\% \\
    & 0 & 0.2 & 1/3 & 72.3\% & 14.6\% & 35.5\% \\
    & 0 & 0.5 & 1/3 & 74.9\% & 12.2\% & 47.5\% \\ \midrule
    \multirow{4}{*}{\textbf{DA-cAE}} & \textbf{0.01} & \textbf{0.005} & \textbf{1/3} & \textbf{81.0\%} & \textbf{6.1\%} & \textbf{9.6\%} \\
    & 0.1 & 0.005 & 1/3 & 78.0\% & 5.5\% & 9.7\% \\
    & 0.2 & 0.005 & 1/3 & 80.3\% & 4.0\% & 11.1\% \\
    & 0.5 & 0.005 & 1/3 & 78.5\% & 3.2\% & 14.0\% \\
    \bottomrule
    \end{tabular}
    \label{results}
\end{table}

We here focus on the MLP task classifier to discuss the impact of parameters in DA-cAE.
As shown in Table~\ref{results}, we first assessed the baseline AE and cAE with $\lambda_A=\lambda_N=r_N=0$ to train the MLP task classifier.
Then, we evaluated the D-cAE with $\lambda_N \in \{ 0.005, 0.01, 0.2, 0.5 \}$, $\lambda_A=0$ and $r_N=1/3$. 
Finally, we fixed $\lambda_N=0.005$ to evaluate our DA-cAE with different choices of $\lambda_A \in \{0.01, 0.1, 0.2, 0.5 \}$.
For each of these parameter combinations, Table~\ref{results} shows the average accuracies of the MLP task classifier (i.e., $4$-class stress level decoding), as well as the accuracy of adversary and nuisance networks (i.e., $20$-class ID decoding).
A higher accuracy of the main MLP task classifier indicates better discrimination of stress levels, a higher accuracy of nuisance network implies that more subject-dependent features are inherent in the representation $z_n$, and a lower accuracy of adversary network demonstrates that subject-invariant task-specific information are preserved in the learned representation $z_a$.
We notice that with increasing $\lambda_N$, the nuisance network accuracy grows and specifically $\lambda_N=0.005$ with $r_N=1/3$ leads to higher task classification accuracy.
Moreover, when fixing $\lambda_N=0.005$ and $r_N=1/3$, we observe that higher $\lambda_A$ censors the encoder with decreased adversary network accuracy, and therefore enforces stronger task-specific information but less extraction of subject-relevant information into the learned $z_a$.

\begin{figure}[t]
  \centering
  \includegraphics[clip,trim=0.8cm 0 0 0.9cm, width=0.47\textwidth]{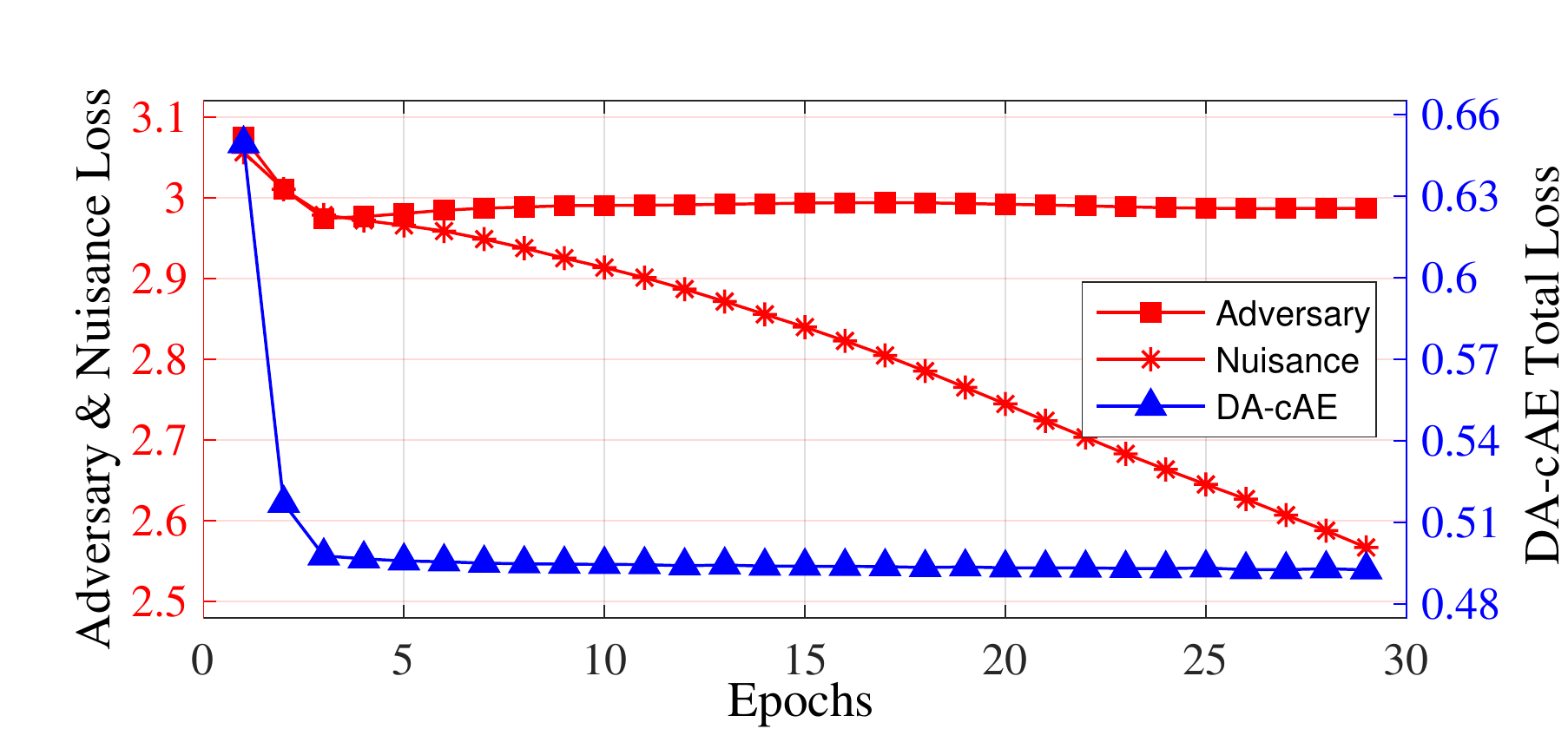}%\vspace{-0.2cm}
  \caption{Convergence of DA-cAE ($\lambda_A=0.01$ and $\lambda_N=0.005$).}
\label{convergence}
\end{figure}

The convergence curve of the optimized DA-cAE case from Table~\ref{results} is shown in Fig~\ref{convergence}, where DA-cAE loss (\ref{DA-cAEloss}) converges within 5 epochs.
With more training epochs, the loss value of the nuisance unit declines steadily, while the adversary loss remains stable as a result of the adversarial relationship between the DA-cAE and adversary classifier, which keeps concealing subject-related information while not disabling the discriminative ability of overall network. 

\begin{figure}[t]
  \centering
  \includegraphics[clip,trim=0.3cm 0 0.8cm 0.1cm, width=\linewidth]{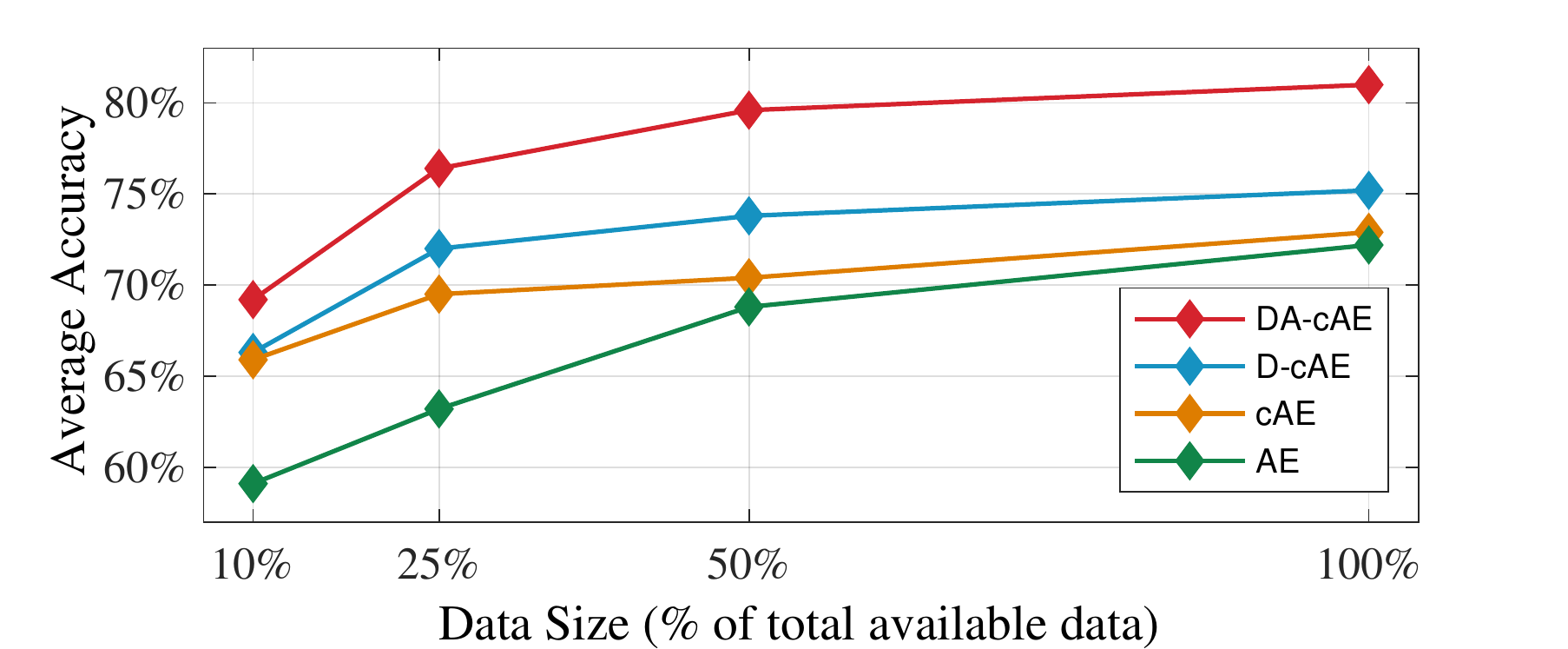}%\vspace{-0.1cm}
  \caption{Classification accuracies with different training dataset sizes, using the optimized model choices of Table~\ref{results}.} % (DA-cAE: $\lambda_A=0.01$, $\lambda_N=0.005$; D-cAE: $\lambda_A=0$, $\lambda_N=0.005$; cAE and AE: $\lambda_A=\lambda_N=0$).}
\label{datasize}
\end{figure}

Finally, we evaluate the impact of training data size on the classification accuracy in Fig~\ref{datasize}. %, where we observed that convergences are achieved via more training epochs with less data.
Regardless of the data size reduction in the available training set, DA-cAE outperforms all models and shows robustness to physiological data size deficiency.
Moreover, it is expected to achieve even higher gain when more measurement data are available for training.
%Overall, we observe that using both adversary and nuisance networks attached to the encoder, the classifier improves the accuracy substantially and shows more stable performance across different left-out subjects.

%%%%%%%%%%%%%%%%%%%%%%%%%%%%%%%%%%%%%%%%%%%%%%%%%%%%%%%%%%%%%%%%%%%%
\section{Conclusion}
\label{sec:conclusion}

We proposed a transfer learning method based on a disentangled adversarial AE model to extract nuisance-robust universal representations from physiological biosignals.
To control the trade-off between task-related features and person-specific information, additional adversary and nuisance networks are jointly trained, such that the feature extractor is applicable to a broader range of users.
We performed a cross-subject transfer analysis based on a public dataset for stress level assessment.
Results showed remarkable benefits of the proposed method in improving both average and worst-case accuracies, indicating better adaptability to new subjects.
Furthermore, extracted features showed universal robustness over different task-specific classifiers. 
We note that the proposed method is applicable to various other data analysis problems.



\begin{thebibliography}{10}
\providecommand{\url}[1]{#1}
\csname url@samestyle\endcsname
\providecommand{\newblock}{\relax}
\providecommand{\bibinfo}[2]{#2}
\providecommand{\BIBentrySTDinterwordspacing}{\spaceskip=0pt\relax}
\providecommand{\BIBentryALTinterwordstretchfactor}{4}
\providecommand{\BIBentryALTinterwordspacing}{\spaceskip=\fontdimen2\font plus
\BIBentryALTinterwordstretchfactor\fontdimen3\font minus
  \fontdimen4\font\relax}
\providecommand{\BIBforeignlanguage}[2]{{%
\expandafter\ifx\csname l@#1\endcsname\relax
\typeout{** WARNING: IEEEtran.bst: No hyphenation pattern has been}%
\typeout{** loaded for the language `#1'. Using the pattern for}%
\typeout{** the default language instead.}%
\else
\language=\csname l@#1\endcsname
\fi
#2}}
\providecommand{\BIBdecl}{\relax}
\BIBdecl

\bibitem{physiological1-HCI}
S.~Jerritta, M.~Murugappan, R.~Nagarajan, and K.~Wan, ``Physiological signals
  based human emotion recognition: a review,'' in \emph{2011 IEEE 7th
  International Colloquium on Signal Processing and its Applications}.\hskip
  1em plus 0.5em minus 0.4em\relax IEEE, 2011, pp. 410--415.

\bibitem{petrantonakis2009emotion}
P.~C. Petrantonakis and L.~J. Hadjileontiadis, ``Emotion recognition from
  \uppercase{EEG} using higher order crossings,'' \emph{IEEE Transactions on
  Information Technology in Biomedicine}, vol.~14, no.~2, pp. 186--197, 2009.

\bibitem{EMG-HANDS-CSL}
M.~Han, S.~Y. G{\"u}nay, G.~Schirner, T.~Pad{\i}r, and D.~Erdo{\u{g}}mu{\c{s}},
  ``\uppercase{HANDS}: a multimodal dataset for modeling toward human grasp
  intent inference in prosthetic hands,'' \emph{Intelligent Service Robotics},
  vol.~13, no.~1, pp. 179--185, 2020.

\bibitem{NonEEG1}
J.~Birjandtalab, D.~Cogan, M.~B. Pouyan, and M.~Nourani, ``A
  non-\uppercase{EEG} biosignals dataset for assessment and visualization of
  neurological status,'' in \emph{IEEE International Workshop on Signal
  Processing Systems}, 2016, pp. 110--114.

\bibitem{NonEEG2}
A.~M. Amiri, M.~Abtahi, A.~Rabasco, M.~Armey, and K.~Mankodiya, ``Emotional
  reactivity monitoring using electrodermal activity analysis in individuals
  with suicidal behaviors,'' in \emph{10th International Symposium on Medical
  Information and Communication Technology}, 2016, pp. 1--5.

\bibitem{NonEEG3}
D.~Cogan, M.~B. Pouyan, M.~Nourani, and J.~Harvey, ``A wrist-worn biosensor
  system for assessment of neurological status,'' in \emph{36th Annual
  International Conference of the IEEE Engineering in Medicine and Biology
  Society}, 2014, pp. 5748--5751.

\bibitem{giakoumis2013subject}
D.~Giakoumis, D.~Tzovaras, and G.~Hassapis, ``Subject-dependent biosignal
  features for increased accuracy in psychological stress detection,''
  \emph{International Journal of Human-Computer Studies}, vol.~71, no.~4, pp.
  425--439, 2013.

\bibitem{giannakakis2019review}
G.~Giannakakis, D.~Grigoriadis, K.~Giannakaki, O.~Simantiraki, A.~Roniotis, and
  M.~Tsiknakis, ``Review on psychological stress detection using biosignals,''
  \emph{IEEE Transactions on Affective Computing}, 2019.

\bibitem{ozdenizci2018time}
O.~{\"O}zdenizci \emph{et~al.}, ``Time-series prediction of proximal aggression
  onset in minimally-verbal youth with autism spectrum disorder using
  physiological biosignals,'' in \emph{40th Annual International Conference of
  the IEEE Engineering in Medicine and Biology Society}, 2018, pp. 5745--5748.

\bibitem{trans2}
H.~Morioka, A.~Kanemura, J.-i. Hirayama, M.~Shikauchi, T.~Ogawa, S.~Ikeda,
  M.~Kawanabe, and S.~Ishii, ``Learning a common dictionary for
  subject-transfer decoding with resting calibration,'' \emph{NeuroImage}, vol.
  111, pp. 167--178, 2015.

\bibitem{trans3}
W.~Tu and S.~Sun, ``A subject transfer framework for \uppercase{EEG}
  classification,'' \emph{Neurocomputing}, vol.~82, pp. 109--116, 2012.

\bibitem{trans4}
S.~Fazli, F.~Popescu, M.~Dan{\'o}czy, B.~Blankertz, K.-R. M{\"u}ller, and
  C.~Grozea, ``Subject-independent mental state classification in single
  trials,'' \emph{Neural Networks}, vol.~22, no.~9, pp. 1305--1312, 2009.

\bibitem{ozdenizci2020learning}
O.~{\"O}zdenizci, Y.~Wang, T.~Koike-Akino, and D.~Erdo{\u{g}}mu{\c{s}},
  ``Learning invariant representations from \uppercase{EEG} via adversarial
  inference,'' \emph{IEEE Access}, vol.~8, pp. 27\,074--27\,085, 2020.

\bibitem{advTrans-csl2}
------, ``Transfer learning in brain-computer interfaces with adversarial
  variational autoencoders,'' in \emph{2019 9th International IEEE/EMBS
  Conference on Neural Engineering (NER)}.\hskip 1em plus 0.5em minus
  0.4em\relax IEEE, 2019, pp. 207--210.

\bibitem{advTrans-csl1}
------, ``Adversarial deep learning in \uppercase{EEG} biometrics,'' \emph{IEEE
  Signal Processing Letters}, vol.~26, no.~5, pp. 710--714, 2019.

\bibitem{advTrans4}
G.~Lample, N.~Zeghidour, N.~Usunier, A.~Bordes, L.~Denoyer, and M.~Ranzato,
  ``Fader networks: Manipulating images by sliding attributes,'' in
  \emph{Advances in Neural Information Processing Systems}, 2017, pp.
  5967--5976.

\bibitem{adversarialClsf}
M.~Han, O.~{\"O}zdenizci, Y.~Wang, T.~Koike-Akino, and D.~Erdo{\u{g}}mu{\c{s}},
  ``Disentangled adversarial transfer learning for physiological biosignals,''
  in \emph{42nd Annual International Conference of the IEEE Engineering in
  Medicine and Biology Society (EMBC)}, 2020.

\bibitem{wu2020modality}
F.~Wu, X.-Y. Jing, Z.~Wu, Y.~Ji, X.~Dong, X.~Luo, Q.~Huang, and R.~Wang,
  ``Modality-specific and shared generative adversarial network for cross-modal
  retrieval,'' \emph{Pattern Recognition}, p. 107335, 2020.

\bibitem{sun2020adversarial}
Y.~Sun, X.-Y. Jing, F.~Wu, J.~Li, D.~Xing, H.~Chen, and Y.~Sun, ``Adversarial
  learning for cross-project semi-supervised defect prediction,'' \emph{IEEE
  Access}, vol.~8, pp. 32\,674--32\,687, 2020.

\bibitem{atzori2016deep}
M.~Atzori, M.~Cognolato, and H.~M{\"u}ller, ``Deep learning with convolutional
  neural networks applied to electromyography data: A resource for the
  classification of movements for prosthetic hands,'' \emph{Frontiers in
  Neurorobotics}, vol.~10, p.~9, 2016.

\bibitem{faust2018deep}
O.~Faust, Y.~Hagiwara, T.~J. Hong, O.~S. Lih, and U.~R. Acharya, ``Deep
  learning for healthcare applications based on physiological signals: A
  review,'' \emph{Computer Methods and Programs in Biomedicine}, vol. 161, pp.
  1--13, 2018.

\end{thebibliography}
\end{document}